\pdfoutput=1

\documentclass[11pt]{article}

\usepackage[final]{acl}

\usepackage{times}
\usepackage{latexsym}

\usepackage{tabularx}
\usepackage{booktabs}

\usepackage[T1]{fontenc}

\usepackage[utf8]{inputenc}

\usepackage{microtype}

\usepackage{inconsolata}

\usepackage{graphicx}

\usepackage{algorithm,xcolor}
\usepackage{algpseudocode}
\usepackage{amsmath}
\usepackage{multirow}
\usepackage{booktabs}
\usepackage{tabularx}
\usepackage{todonotes}

\title{Small Models, Big Results: Achieving Superior Intent Extraction through Decomposition}

\author{
 \textbf{Danielle Cohen\thanks{These authors contributed equally.}\textsuperscript{1}},
 \textbf{Yoni Halpern$^*$\textsuperscript{1}},
 \textbf{Noam Kahlon\textsuperscript{1}},
 \textbf{Joel Oren\textsuperscript{1}},
\\
 \textbf{Omri Berkovitch\textsuperscript{1}},
 \textbf{Sapir Caduri\textsuperscript{1}},
 \textbf{Ido Dagan\thanks{co-senior author.}\textsuperscript{1,2}},
 \textbf{Anatoly Efros$^\dagger$\textsuperscript{1}}
\\
\\
 \textsuperscript{1}Google,
 \textsuperscript{2}Bar-Ilan University
\\
 \small{
    \textbf{Correspondence:} \href{mailto:daniellecn@domain}{daniellecn@google.com}, \href{mailto:yhalpern@domain}{yhalpern@google.com}
 }
}

\begin{document}
\maketitle
\begin{abstract}
Understanding user intents from UI interaction trajectories remains a challenging, yet crucial, frontier in intelligent agent development. 
While massive, datacenter-based, multi-modal large language models (MLLMs) possess greater capacity to handle the complexities of such sequences, smaller models which can run on-device to provide a privacy-preserving, low-cost, and low-latency user experience, struggle with accurate intent inference. 
We address these limitations by introducing a novel decomposed approach: first, we perform structured interaction summarization, capturing key information from each user action. Second, we perform intent extraction using a fine-tuned model operating on the aggregated summaries. 
This method improves intent understanding in resource-constrained models, even surpassing the base performance of large MLLMs. 
\end{abstract}

\section{Introduction}

Advancements in the capabilities of multi-modal large language models (MLLMs) has led to recent interest in modeling sequences of user interactions with phone and web graphical interfaces, both for the purposes of automation~\citep{wang2024gui,martinez2024screenshot}, and understanding~\citep{berkovitch2024identifying, zhang2024summact}.

In this work, we focus on the user intent extraction task, which consists of producing a free-form description of the inferred intent of a user from a sequence of interactions with a device.

Large MLLMs are naturally fairly good at this task, however, it is more challenging for smaller models (E.g., Gemini 1.5 Flash 8B~\citep{gemini1_5} or Qwen2 VL 7B~\citep{Qwen2VL}). The performance of smaller models is important for user interaction tasks due to their ability to operate within a private, on-device environment like a phone or browser, with reduced cost, energy usage, and latency~\citep{xu2024device}.

\begin{figure*}[t]
    \centering
    \includegraphics[width=1.0\textwidth]{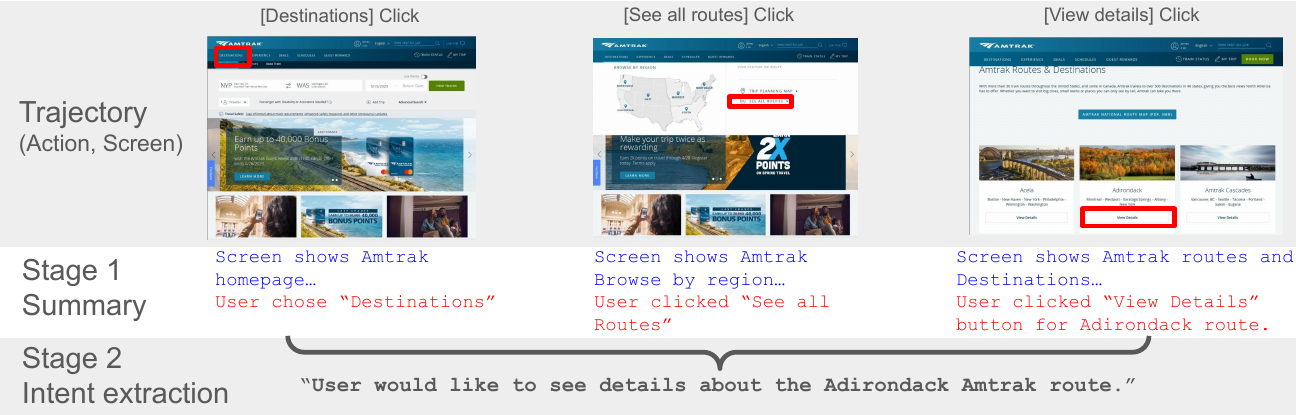}
    \caption{Proposed intent extraction flow (Described in detail in Section~\ref{sec:method}). Individual interactions (input as action strings and screenshots) are summarized and then the summaries are combined to output a short inferred intent for the trajectory. The summarization step uses both action strings and visual screen information to output a structured summary with two fields corresponding to screen summary (top, blue) and user action (bottom, red).}
    \label{fig:flow}
\end{figure*}

In this paper, we introduce a two-stage approach for extracting user intent with small models. In the first stage, each atomic interaction is summarized. In the second stage, the full sequence of summaries is fed to a second model which outputs an intent. The overall flow is illustrated in Figure~\ref{fig:flow}. Using semantic equivalence metrics on public UI automation data, our two-stage approach demonstrates superior performance compared to both smaller models and a state-of-the-art large MLLM, independent of dataset and model type. Our approach also naturally handles scenarios with noisy data that traditional supervised fine-tuning methods struggle with. The modular nature of the architecture is helpful from an engineering perspective, allowing us to evaluate the approach in detail and identify key areas to improve. 

Our contributions can be summarized as follows: 
1. We describe an effective decomposition of intent-extraction that unlocks the potential of small models; 
2. We present non-trivial design components related to each stage of the decomposition; 
3. We extensively evaluate our approach and demonstrate the effectiveness of our method across a range of data sets, base models and metrics.

\section{Background}
\subsection{Intent Extraction from UI Interactions}

We formalize the intent extraction task, sometimes called goal understanding, similarly to prior works ~\citet{berkovitch2024identifying} and \citet{zhang2024summact}.
Consider a user journey $T$ within a mobile or web application, represented as a sequence of interactions: $T=(I_1, I_2, ..., I_n)$, where each interaction $I_i=(O_i, A_i)$ consists of an observation, $O_i$, and the action, $A_i$, the user performed at that step. This description is general and different modeling approaches have used different representations for observations and actions (e.g., textual descriptions, screenshot images, DOM hierarchies, etc.) \citep{rawles2023android, burns2022dataset}. 
The objective of the intent-extraction task is to generate a free-form sentence describing the user's intent. Effectively, this setting can be thought of as the inverse problem of the UI automation task, with inputs and outputs swapped. Rather than producing a sequence of actions from an instruction, we ask ``what was the user trying to accomplish with this trajectory?''. Intent extraction has been identified as an important building block for UI automation tasks, proactive assistance, and personalized memory~\citep{berkovitch2024identifying, zhang2024summact}.

Very recently, a few works have begun addressing intent extraction from UI interactions. \citet{berkovitch2024identifying} proposed this novel task, and seemingly were the first to point out that it can be viewed as the inverse task of UI Automation. They evaluated MLLM intent-extraction performance over UI automation datasets (swapping input/output roles). As input, they considered screenshot images and textual descriptions of user actions, as we do in our work, and assessed the performance of standard MLLMs using a fairly simple prompting approach. In our work, we follow this evaluation approach, while testing an improved version of their prompt as a baseline. 
\citet{zhang2024summact} also evaluated their SummAct model over UI automation datasets, but took as input \textit{only} short textual descriptions of the specific UI elements with which the user interacted and the respective user actions, without considering the global screen context, as we do. Loosely similarly to our method, they decomposed intent generation into two consecutive stages, though these are substantially different than ours and required intervening with the LLMs attention mechanism for fine-tuning (see Appendix \ref{appendix:compwithsummact} for detailed comparison of the two methods).
Finally, the UI-JEPA model \citep{fu2024ui} takes as input videos of the entire UI session. Their method adapts a specialized video-embedding method to work with videos of UI sessions, and then finetunes an LLM decoder that generates intent descriptions based on these embeddings. Overall, SummAct, our work, and UI-JEPA consider different types of inputs, providing increasingly richer contexts with increasing complexity: ``local'' text descriptions of the user action and its respective UI element, accounting for the full screen image, and processing a complete video of the entire session (respectively). 
It is left for future work to thoroughly explore the pros and cons of these alternative inputs under different scenarios.\footnote{An empirical comparison with SummAct on equal grounds was not possible, since their code was released without full prompts close to the submission deadline. Similarly, the UI-JEPA inference code hasn't been released yet while the license on the dataset restricts our lab's usage.}

With respect to evaluating model-predicted intents, a good intent is {\em faithful:} only describes things that actually occur in the trajectory; {\em comprehensive:} provides all of the information about the user intent required to re-enact the trajectory; and {\em relevant:} does not contain extraneous information beyond what is needed for comprehensiveness. However, even with a well-defined ground truth intent, accurately evaluating a model's extracted intent is challenging. User intents often contain many details, such as trip planning specifics or transaction data, which require metrics that can handle partial matches. Such metrics fall into two categories: semantic, which analyze underlying meaning, and lexical, which assess surface-level word overlap. As \citet{caduri2025bifact} show, lexical metrics (e.g., BLEU and ROUGE) correlate poorly with human judgments of intent similarity, as they merely compare words. In contrast, semantic metrics, such as NLI (Natural Language Inference) and BI-Fact (a bi-directional variant of FActScore~\citep{min2023factscore}), strive to capture the intended meaning.

Further, intent extraction is inherently subjective, as a single trajectory could have been driven by multiple underlying motivations (e.g., a user may have selected a flight based on its price versus its departure time). This subjectivity is evident in prior work, such as \citet{berkovitch2024identifying} where 
human-composed intentions matched each other in only 80\% and 76\% of web and phone trajectories, respectively. This level of human agreement may be considered a practical upper bound for performance on this task.

\subsection{UI Interaction Datasets}
\label{sec:background:datasets}

Recently, a number of datasets have been developed for evaluating UI interaction agents, (surveyed in \citet{wang2024gui}). We use two that are representative and suitable for measuring the intent extraction task. We confirmed that our usage of the data adhered to all ethical and legal standards.

\noindent{\bf Mind2Web} (CC BY 4.0 license) \citep{deng2024mind2web}: Has 2,350 human demonstrations on websites. Each user trajectory is on average 7.3 steps long and contains screenshots and actions for each step, as well as a high level description of the task the human was asked to perform. 

\noindent{\bf AndroidControl} (Apache 2.0 license) \citep{li2024effects}: Has 15,283 examples of humans performing tasks on Android apps. Each user trajectory is on average 5.5 steps long, and contains screenshots and actions for each step, as well as a high level description of the goal. 

Mind2Web's data collection included a validation step where annotators verified the alignment between the completed trajectory steps and the intent, making this dataset highly suitable for the intent extraction task as well. This crucial step was absent from the AndroidControl collection protocol, resulting in noisier labels. For example consider the following task ``Delete all emails from sender X'' in a scenario where there were no emails from that sender. Based on the execution of task it is impossible to identify that the original goal was to delete emails. We preprocess labels to remove clearly irrelevant statements~(Section~\ref{sec:datasets_preprocessing}) and analyze the effect of remaining discrepancies between the labels and trajectories in Section~\ref{sec:label_quality}.

\subsection {Related Research Lines}

\paragraph {\bf User interaction understanding for HCI} Single screen summarization as a special case of image description has been extensively studied for the purposes of e.g.,  accessibility, automation, and question answering~\citep[e.g.,][]{li2021screen2vec, bai2021uibert, li2022spotlight, wang2021screen2words, yang2024aria}. Our setting of identifying and summarizing intents from trajectories has been recently proposed in \citet{berkovitch2024identifying, zhang2024summact, martinez2024screenshot}.

\paragraph {\bf Multi-stage summarizations} Decomposing a complex task into smaller simpler stages is a well-known approach for problem solving. 
Hierarchical models are common in summarization tasks of many modalities, e.g., text~\citep{christensen2014hierarchical}, audio~\citep{li2021hierarchical}, video \citep{zhao2022hierarchical, cheng2024enhancing}. Chain of Thought (CoT) reasoning~\citep{wei2022chain} is a popular general-purpose prompting method to decompose a problem into smaller parts. \citet{khot2022decomposed} propose an automated decomposition step that delegates different parts of the problem to distinct model calls.

\section{Baseline Modeling Approaches}
\label{sec:baselines}


In this section, we first present natural baseline approaches for addressing our task, whose lessons led us to develop our decomposed two-stage approach which will be described in Section~\ref{sec:method}. 
Our task is a text generation task, where intent descriptions are generated from the multi-modal input of UI trajectories. As such, it is most natural to address it through multi-modal LMs, applying either prompt-based or fine-tuned methods. The focus of our work is to explore the use of small LMs, for eventual utilization on-device. The particular models used are specified in Section \ref{sec:experimental_setting:models}, including a top-tier large model as a reference point.

\paragraph{Prompt-based methods}
Such methods are advantageous in that they do not require training data, instructing a \textit{generic} LM via its prompt. We found that a CoT prompt worked best. Specifically, our CoT prompt (see \ref{fig:prompt_cot}) instructs the model to first generate a sequence of individual descriptions of the user intents within \textit{each UI interaction}, and then to consolidate these interaction-level description into the final description of the accumulated user intent along the trajectory. 

\paragraph{Fine-tuned models}
Since performance of prompting a generic model may not be fully aligned with the intended task output and prompt-based performance of small LMs might generally be limited,
we also explore baseline fine-tuning methods. To that end, we fine-tuned small models using available training datasets, specifically those developed for the inverse problem of UI automation, while swapping their input/output roles (see Section \ref{sec:background:datasets}). 

Both prompt-based and fine-tuned baselines require  large context window to contain the entire user trajectory including images. As described in Section \ref{sec:experimental_setting:models}, practically this required some filtering over the input to fit the available context size.

\section{A Decomposed Two-stage Model}

\label{sec:method}

While CoT prompting works well with large language models (LLMs), we observe limitations in both CoT and fine-tuned small LMs when presented with the full trajectory. When applying CoT reasoning, small models struggle to generate high-quality thoughts that cover the full trajectory. Fine-tuned small models also have trouble generating comprehensive intents from the full trajectory.

These observations led us to develop a decomposed, two-stage approach that emulates the CoT process, illustrated in Figure \ref{fig:flow}. First, we use prompting to generate a summary for each interaction (consisting of a visual screenshot and textual action representation) in a trajectory. This stage is prompt-based as there is currently no training data available with summary labels for individual interactions. Second, we feed all of the interaction-level summaries into a second stage model  to generate an overall intent description. We apply fine-tuning in the second stage and we describe that process in more detail below (Section~\ref{sec:session_Level_Intent}). 
The following subsections provide a detailed description of each stage in our proposed method.

\subsection{Interaction Summarization}
\label{sec:interaction_summarization}

In the first stage, we summarize each individual user interaction $I_i=\{O_i, A_i\}$ of the length-$n$ trajectory $T = (I_1,\ldots,I_n)$. The summarization uses visual and textual information to extract relevant information regarding the user's goals and actions within that interaction. The output of this stage is a summary of the screen context and user action (see Figure~\ref{fig:flow}). This key information, which describes this particular user interaction, will be used in the subsequent fusion stage. This summarization process is entirely prompt-based (see ~\ref{fig:summarisation_prompt}).

We add the two following improvements to the design of this stage, which improves overall performance, as shown in ablation studies in Section~\ref{sec:ablation}.

\paragraph {\bf Context window} While the primary task is to understand $I_i$ in isolation, we recognize that often context can be crucial for eliminating ambiguity and/or uncertainty. Therefore, in addition to $I_i$ the model also receives as input the preceding and successive interactions, $I_{i-1}$ and $I_{i+1}$, respectively. This allows the model to use e.g., the visual cues from both the current the next screenshot to understand the user action at step $i$. 

\paragraph {\bf Structured summaries} We request that the summary be structured in two distinct components: (a) the relevant screen context -- a short list of salient details on the current screen $O_i$, and (b) the user action: a list of mid-level actions that the user took in the current interaction (example in Figure~\ref{fig:flow}). Despite being structured in two fields, the visual cues from the screenshot are also used to understand the action (e.g., in Figure~\ref{fig:flow}, the visual cues are helpful to extract that the click relates to the ``Adirondack'' section of the page.) As a practical measure for dealing with cases where the model outputs its (unwarranted) interpretations of the user's underlying intent, we also instructed it to output those in a third field (labelled ``\textit{speculative intent}'') that we discard before proceeding to the next stage.

This structured format was selected to address challenges encountered with alternative prompting strategies.  Simply asking the model to be concise resulted in summaries that lacked crucial details.  Conversely, prompting for comprehensive information, including user intent, led to excessive speculation that could hinder the subsequent summary fusion stage. 
Our structured format aims to capture a broader range of information while enabling the removal of speculative elements prior to the second stage. This balanced approach mitigates the risk of contradictions and improves the overall summarization process. 

\subsection{Generating Session-Level Intent}
\label{sec:session_Level_Intent}
In the second stage, we aggregate the information extracted during the first stage. A second-stage model takes as input the summaries of all interactions in the trajectory to infer the user's overall intent. This aggregation stage is implemented by fine-tuning a model to specialize in the aforementioned aggregation. For fine-tuning, the training data consists of: a. input summaries representing all interactions in the trajectory, and b. a corresponding ground truth target that describes the user's overall intent in the given trajectory. For comparison, a variation with no fine-tuning, which is fully prompt-based, is also available as part of our ablation study in Subsection \ref{sec:ablation}.

We noted in early explorations that naively applying fine-tuning yields a model that embellishes or hallucinates by introducing details that were not present in the screen summary inputs. On further examination, we found that the training procedure encourages the model to act this way since the inputs are potentially incomplete summaries and the targets are the complete intent statements. Thus, when looking at (input, target) pairs, the model learns that it needs to sometimes add additional information in order to produce the target intent.

Following this insight we refine our target intents at training time to remove details not reflected in the corresponding input (using a large language model, see Figure~\ref{fig:label_refinement_prompt} for details on the prompt used in this stage). This ensures that the model will learn to infer intents based solely on the provided interaction summaries. We discuss the effects of this cleanup stage in Subsection~\ref{sec:ablation}.

\section{Experimental Setting}
\label{sec:experimental_setting}
\subsection{Models}
\label{sec:experimental_setting:models}
We focus on smaller, multi-modal models, that can be fine-tuned. In particular, we use {\em Gemini\footnote{Terms of service: https://ai.google.dev/gemini-api/terms} 1.5 Flash 8B}~\citep{gemini1_5} and {\em Qwen2 VL 7B} (Apache 2.0 license) ~\citep{Qwen2VL}. For comparisons with a MLLM, we use {\em Gemini 1.5 Pro}~\citep{gemini1_5}.

When using Qwen2 VL 7B for baseline models, we dropped frames randomly from the trajectory if they exceeded the context window length. We found that limiting trajectories to 15 steps was sufficient to run our experiments. We also downsized AndroidControl images by a factor of 4 in each dimension when inputting them to Qwen models. Details of fine-tuning can be found in Appendix~\ref{appendix:finetuning_details}.

\subsection{Datasets and Preprocessing}
\label{sec:datasets_preprocessing}

We use the Mind2Web~\citep{deng2024mind2web} and  AndroidControl~\citep{li2024effects} datasets as representative user interaction datasets. We follow the standard train/test split of each dataset, fine-tuning with train, and reporting results on test data.

In Mind2Web, we represent the action from the dataset textually: (e.g., ``[element name] click'' or ``[element name] hover''.). In AndroidControl, we use the accessibility tree to convert the screen coordinates of the interacted item to an element  name and format the action in the same way. In both datasets, we use screenshots as observations. 
We highlight the interacted element in the screenshot with with a red box~\citep{zhenggpt, yang2023set}. To improve the evaluation of user intent interpretation, goal labels from the datasets were cleaned and restructured to separate platform-specific information from the core intent, see Appendix~\ref{appendix:preprocessing_details} for more details.

\subsection{Evaluation Metrics}
\label{sec:experimental_setup:metrics}
We measure quality of extracted goals using two different semantic equivalence metrics.

\paragraph {\bf T5 NLI}~\citep{honovich-etal-2022-true-evaluating}: A T5-XXL model\footnote{Available at: \url{https://huggingface.co/google/t5_xxl_true_nli_mixture}} trained for NLI (Natural Language Inference). We compute the entailment probability of the produced summary from the gold standard and vice versa, and then average the two values to get a single bidirectional score.

\paragraph {\bf BiFact}~\citep{caduri2025bifact}: A bidirectional variation of FActScore~\citep{min2023factscore} developed for assessing the equivalence of intents in UI interactions, demonstrating the highest correlation with human judgments compared to existing methods. This metric deconstructs both the ground-truth and predicted intents into their fundamental factual components using an LLM (we use Gemini 1.5 Pro for this). These components are then compared to measure the extent of coverage. We use the BiFact measures of precision (the proportion of facts in the predicted intent that are present in the true intent), recall (the proportion of facts in the true intent that are captured by the predicted intent) and F1. To assess the robustness of our method, we applied the BiFact metric with an alternative model not used in our experiments. The results remained consistent (Appendix \ref{appendix:alternative_llm}).

We believe that BiFact, which uses a fine-grained, fact-level comparison, is ideally suited for our task since intents can be composed of many parts (e.g., book a flight, flight is to LAX, flight is on Friday).  NLI, which holistically evaluates logical entailment of the full sentences is less ideal, but provides an extra signal.

\section{Experiments}
\begin{table*}[htbp]
    \centering
    {\small
    \begin{tabularx}{\textwidth}{l|XXX|X|XXX|X}
\toprule

& \multicolumn{4}{c}{\textbf{Mind2Web}} & \multicolumn{4}{c}{\textbf{AndroidControl}} \\
\midrule
& \multicolumn{3}{c|}{BiFact} & \multicolumn{1}{c|}{Bi-NLI} & \multicolumn{3}{c|}{BiFact}  & \multicolumn{1}{c}{Bi-NLI} \\
Method & F1 & Precision & Recall &  & F1 & Precision & Recall &  \\
\midrule
Gemini Flash 8B  \\
\midrule

CoT & 0.659 & 0.758 & 0.647 & 0.326 & 0.594 & 0.628 & 0.660 & 0.302 \\
E2E-FT & 0.653 & 0.676 & 0.671 & 0.311 & 0.611 & 0.655 & 0.656 & 0.343 \\
Decomposed-FT & \textbf{0.752} & \textbf{0.814} & \textbf{0.746} & \textbf{0.391} & \textbf{0.630} & \textbf{0.664} & \textbf{0.688} & \textbf{0.350} \\
\midrule
Qwen2 VL 7B \\
\midrule
CoT & 0.563 & 0.694 & 0.551 & 0.272 & 0.538 & 0.589 & 0.603 & 0.280 \\
E2E-FT & 0.610 & 0.670 & \textbf{0.621} & 0.233 & 0.506 & 0.594 & 0.546 & \textbf{0.343} \\
Decomposed-FT & {\bf 0.623} & {\bf 0.736} & 0.609 & \textbf{0.300} & {\bf 0.608} & {\bf 0.661} & {\bf 0.646} & 0.333\\
\midrule 
Gemini-1.5-Pro  \\
\midrule
CoT & 0.730 & 0.773 & 0.745 & 0.331 & 0.634 & 0.612 & 0.767 & 0.347 \\
\bottomrule
\end{tabularx}
}
    \caption{BiFact and Bi-NLI results on the Mind2Web and AndroidControl datasets using Gemini 1.5 Flash 8B, Qwen2 VL 7B, and Gemini 1.5 Pro. Best scoring method for each model is bolded. F1, precision, recall are micro-averaged over the dataset.}
    \label{tab:gemini_mind2web}
\end{table*}

\subsection{Evaluating Extracted Intents}

\normalsize
To show that our decomposed approach is generally helpful compared to baselines across models and data modalities, we evaluate the metrics in Section~\ref{sec:experimental_setup:metrics} on two different datasets using two different models. The results are displayed in Table~\ref{tab:gemini_mind2web}. 

In this table, CoT (Chain of Thought) and E2E-FT (End-to-End fine-tuned) represent the natural baselines described in Section~\ref{sec:baselines}. Of these two baselines, neither is uniformly more effective across all settings. 
On the Mind2Web dataset, which has cleaner labels (described in~\ref{sec:background:datasets}), Gemini, as a stronger base model, has higher BiFact F1 and Bi-NLI scores with CoT, whereas Qwen2 VL 7B benefits from fine-tuning. 
Gemini 1.5 Pro CoT is presented as a comparison to a top-tier large MLLM. We find that on Mind2Web, the fine-tuned decomposed approach allows the Gemini Flash 8B to even exceed the performance of the Gemini 1.5 Pro model using CoT. On AndroidControl, the scores are comparable. 

The BiFact score is non-deterministic as it uses an LLM to compute the score. We observe a 0.016 standard deviation on repetition. A more detailed breakdown of performances on the test sets by heldout data type can be found in Appendix~\ref{appendix:test_set_breakdown}.

\noindent {\bf Manual verification - Human preference}: To further verify, a human rater compared 20 Mind2Web trajectories with intent predictions from Gemini Flash 8B, choosing between CoT and Decomposed-FT responses (details in Appendix~\ref{appendix:human_preference}). Overall, Decomposed-FT was preferred in 12 instances, CoT in 4, and 4 were rated equally.

\subsection{Label Quality and Comparison with Expert-Written Intents}
\label{sec:label_quality}

To understand the quality of the labels after preprocessing (described in Section~\ref{sec:datasets_preprocessing} and Appendix~\ref{appendix:preprocessing_details}), we elicited expert-written intent statements for 100 examples in the AndroidControl dataset following the annotation protocol in~\citep{berkovitch2024identifying}. In Table~\ref{tab:expert_labels} we compare the BiFact F1 metric for proposed intents against dataset labels and against expert written intents (more detailed metrics in Appendix Table~\ref{tab:detailed_annotated_vs_gold}).

Overall, the performance of each model improves when compared to expert annotations, except for the E2E-FT model, which was trained on the noisy labels.  The fine-tuned decomposed approach also uses fine-tuning, and could have been expected to similarly suffer from training on noisy labels, but instead it
significantly improves when evaluated using expert intents. We believe this is due to our approach to constructing fine-tuning labels (Section~\ref{sec:session_Level_Intent}) which removes information present in the gold labels but absent from summaries. Interesting to notice that after cleaning the AndroidControl data, the performance of Gemini 1.5 Pro CoT is similar to the performance on Mind2Web suggesting the gap in performance between datasets is mainly the result of data noise.

\begin{table}[htbp]
    \centering
    {\small 
    \begin{tabular}{p{3.5cm}p{1.5cm}p{1.5cm}}
\toprule
Method\newline (Gemini-1.5 Flash 8B) & Expert\newline Labels & Dataset\newline Labels \\
\midrule
CoT & 0.652 & 0.580 \\
E2E-FT & 0.590 & 0.565 \\
Decomposed-FT & {\bf 0.701} & {\bf 0.596} \\
\midrule
Gemini-1.5-Pro CoT & 0.724 & 0.635 \\ 
\bottomrule
\end{tabular}
    \caption{A comparison of BiFact F1 scores for intent prediction on the AndroidControl dataset, using expert annotations and dataset labels as ground truth. A more detailed table appears in Appendix~\ref{appendix:expert_labeling}.}
    \label{tab:expert_labels}
    }
\end{table}

\subsection{Ablation Study}

\label{sec:ablation}
We consider four variants of our method to estimate the impact of each design choice. The performance of each of these ablations can be found in Table~\ref{tab:ablation}. 

\paragraph{No Context} In this variant, Stage 1 is provided with only a single interaction, without previous or next interactions. Our analysis reveals that incorporating information from the previous and next interactions significantly helps the model to infer the user action in the current screen, thereby leading to a noticeable increase in Stage 1 recall. 

\paragraph{Unstructured Interaction-level Summaries}  
Our method instructs the model to output interaction summaries that are structurally broken down into context, user actions, and a speculative intent list (which is removed prior to proceeding to the next stage). Instead, we permit free-form summaries, and the concatenation of those are provided to the goal extraction. Instructing the model to output these particular structured responses allows the Stage 2 model to focus on user actions on the one hand, while mitigating Stage 1 hallucinations as much possible. We notice a slight decrease in both precision and recall, as a result of eliminating this part in our method.

\paragraph{No Fine Tuning}
In this ablation, the second stage of our model was not subjected to fine-tuning and operated solely on a prompt-based approach. Our findings indicate that this configuration led to a marked decrease in precision. Without fine-tuning, the model tended to be more verbose, resulting in a higher proportion of irrelevant or incorrect information being generated. Conversely, this same verbosity contributed to an increase in recall, as a broader range of potential information was captured. However, when considering the F1 score, which balances precision and recall, the fine-tuned version of Stage 2 demonstrated superior performance, underscoring the benefits of the fine-tuning process. For completeness, an analysis of this prompt-based approach on larger models is provided in Appendix~\ref{appendix:test_decomposed_analysis}.

\paragraph{No Label Refinement} Recall that label refinement was added to address Stage 2 hallucinations. In this variant, we exclude the label refinement step, during the data preparation for the fine-tuning of the Stage 2 model,  as described in Subsection~\ref{sec:session_Level_Intent}. As expected, after removing this step, we notice a significant decrease in precision. However, we also see a slight increase in recall, suggesting potential areas for improvement in the refinement process.

\begin{figure*}[h]
    \centering
    \includegraphics[width=0.9\textwidth]{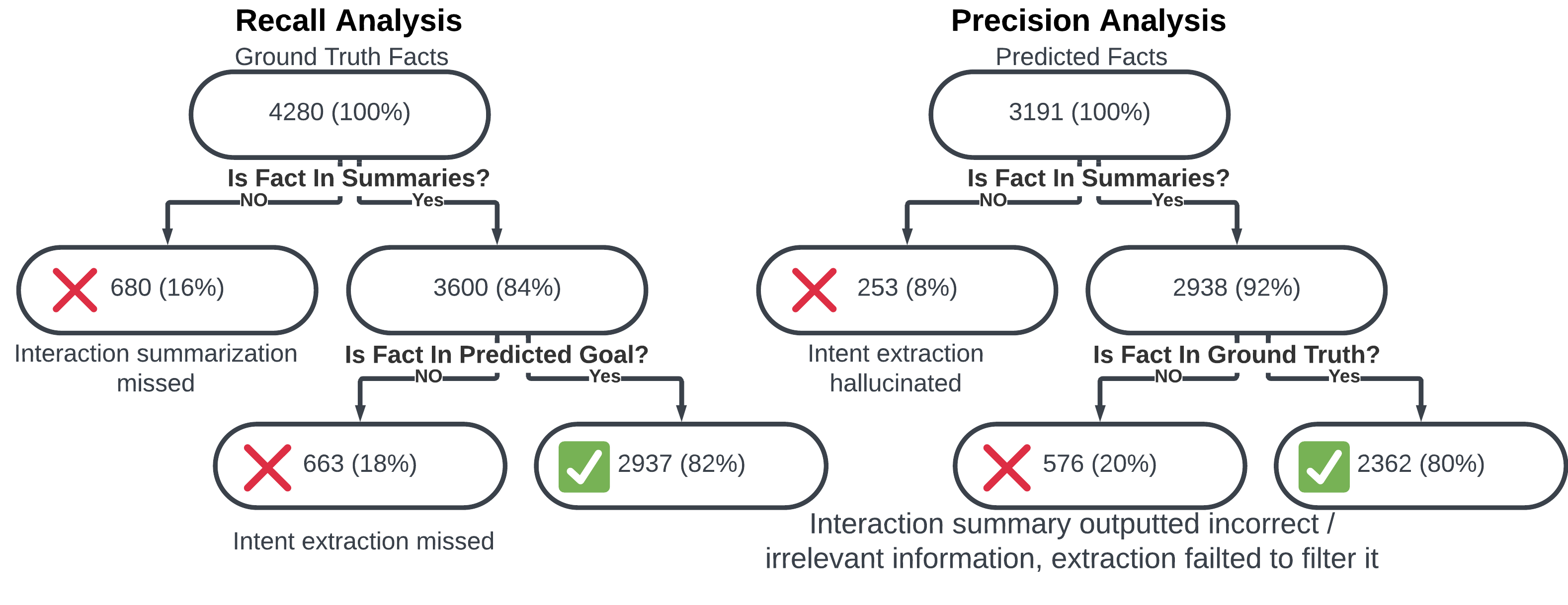}
    \caption{Error propagation analysis of our method on the Mind2Web dataset, tested with Gemini Flash-8B, tracking ground-truth and predicted facts to obtain stage-level recall and precision.}
    \label{fig:funnel}
\end{figure*}

\begin{table}[!ht]
    \centering
    {\small
    \begin{tabular}{lccc}
    
\toprule
Method & F1 & Precision & Recall \\
\midrule
Decomposed-FT & {\bf 0.752} & {\bf 0.814} & 0.746 \\
- No context & 0.711 & 0.794 & 0.698 \\
- Unstructured & 0.731 & 0.791 & 0.737 \\
- No fine-tuning & 0.724 & 0.719 & {\bf0.802} \\
- No label refine & 0.738 & 0.756 &  0.773 \\
\bottomrule

\end{tabular}
}
\caption{Ablation study on Mind2Web using BiFact scores. The Decomposed-FT model is the full model and then each subsequent line shows the effect of removing a single design component.}
\label{tab:ablation}
\end{table}

\subsection{Manual Error Analysis}

To gain a deeper understanding of the errors produced by the decomposed-FT model, we manually analyzed 20 examples.

Counts are indicated in parentheses after each error type. Some examples exhibited multiple error types, so the counts do not necessarily add up to the total number of examples.

\paragraph{Incorrect screen understanding (6)}  Includes instances where the model misinterpreted the UI elements or incorrectly understood the user action.
\paragraph{Summary omissions (6)} Includes instances where the model failed to capture important on-screen details, like omitting the destination on a travel site.
\paragraph{Hallucinations (4)} Includes instances involving generating information not present on the screen, such as claiming the user selected a specific item when they did not.
\paragraph{Irrelevant details (0)} Includes instances where the model included correct but excessive information. While this error was not present in our full model, it was significant in the "no formatting" models used in the ablation study (Section~\ref{sec:ablation}).

\paragraph{Intent extraction omissions (8)} Includes instances where the second stage failed to include important details present in the individual summaries.

\paragraph{Evaluation Errors (1)} These errors were infrequent and typically involved situations where complex screen understanding was required to determine the equivalence of intents. 

The majority of issues occur in the Interaction Summarization stage, suggesting potential benefits from distillation training of this stage. Initial experiments, showed no significant improvements from distillation. Further investigation is warranted.

\subsection{Error Propagation Analysis}
\label{sec:performance}

Obtaining a correct intent from the Decomposed-FT method requires the two stages in Section~\ref{sec:method} to work together effectively. In this section, we investigate error propagation through the two stages using the BiFact decomposition of intents and summaries into atomic facts.

Missed facts, resulting in lowered recall, can occur when a fact is missing in the interaction summarization stage (interaction summarization miss) or the fact can be present in the first stage, but incorrectly dropped in the intent extraction phase (intent extraction miss). An irrelevant or incorrect fact, resulting in lowered precision, can be introduced in the interaction summarization phase and propagated through intent extraction (summarization introduced), or it can be absent from the interaction summarization phase and introduced in the intent extraction phase (intent extraction hallucinated).

Our analysis of the Mind2Web test set is given in Figure~\ref{fig:funnel} using the Decomposed-FT model. The left-hand side, which focuses on recall, shows that the summarization process results in a 16\% loss of ground truth facts. Subsequently, intent extraction further reduces the remaining facts by 18\%. Effectively, each stage introduces a similar magnitude of error. 
The right-hand side describes the precision analysis, showing that 8\% of the facts predicted by Decomposed-FT were, in fact, hallucinations. This low hallucination rate is attributed to the label processing techniques employed during training. Following that, 20\% of the remaining predicted facts were present in the summary but absent from the ground truth, indicating incorrect or irrelevant information in the interaction summarization output and a filtering issue of the intent extractor. We propose this analysis framework to evaluate future two-stage intent extraction methods, aiming to optimize future efforts and assess each stage's impact on overall performance.

\section{Computational Cost and Latency Analysis}
\begin{table*}[htbp]
\centering
\small 
\setlength{\tabcolsep}{4pt} 
\begin{tabular}{lcccc}
\hline
\textbf{Model} & \textbf{\shortstack{Input \\ tokens}} & \textbf{\shortstack{Output \\ tokens}} & \textbf{\shortstack{Price per million \\ USD (eq.\ref{eq:1})}} & \textbf{\shortstack{Latency \\ (eq.\ref{eq:2})}} \\ \hline
E2E & 1839 & 20 & 191.9 & 0.24 \\
CoT & 1961 & 127 & 246.9 & 0.43 \\
Decomposed FT & 2103 & 622 & 600 & 0.6 \\
\shortstack{Decomposed FT (Latency-optimized)} & 2009 & 514 & 406.5 & 0.24 \\ \hline
\end{tabular}
\caption{A summary of computational cost and latency}
\label{tab:computation_latency_condensed}
\end{table*}

We conducted a comparative analysis of expected computational cost and latency, for our method vs. the baselines. To provide reliable figures, we relied on published performance benchmarks for the runtime of LLMs, from which expected runtime can be estimated as a function of the number of calls and the number of input and output tokens. We prefer this calculation over directly measuring runtime in our computational environment, since it does not provide easy means to isolate exact computation times from confounding factors such as network traffic and resource contention. We report computational cost based on pricing of the gemini developer API and expected latency using independent data from https://artificialanalysis.ai (as of June 2025).
\paragraph{Part 1. Estimated total computational cost} This is the estimated cost for generating an intent description, summing over all involved LLM calls.

\begin{equation}
\begin{split}
\text{Price per million (USD)} = {}& 0.1 \times \text{input tokens} \\
                                  & + 0.4 \times \text{output tokens}
\end{split}
\label{eq:1}
\end{equation}

\paragraph{Part 2. Latency from the last user interaction} until the intent description is available for further processing (e.g. for generating follow up suggestions or automation). The latency computation assumes that all screen summaries are generated along the user session, at the end of each interaction step. Thus, end-of-session latency involves only generating the summary of the last screen and the call for generating the intent description from all screen summaries.

We further measure latency for a latency-optimized decomposed variant. In this variant, instead of summarizing the final screen, the visual input is fed directly to the intent generation along with the summaries of all preceding screens, thus incurring latency only for the intent generation LLM call. This variant does not lose quality compared to the original Decomposed FT model.

\begin{equation}
\begin{split}
\text{Total latency } 
    &\approx \text{Time To First Answer Token} \\
    &\quad + \frac{\text{num\_output\_tokens}}{\text{Output Tokens per Second}} \\
    &\approx 0.2 + \frac{1}{550} \times \text{num\_output\_tokens}
\end{split}
\label{eq:2}
\end{equation}

\noindent The size of the input appears to be a negligible factor when the input is between 100-1K tokens \footnote{\url{https://artificialanalysis.ai} Latency and Output Speed by Input Token Count Context Length}. Images are estimated as a fixed 256 tokens each. Overall, the gain in performance from the Decomposed FT approach does come with an additional 2-3x computational cost over the small model baselines that we quantified here. The original Decomposed FT approach would be expected to add latency, which could be an issue in latency sensitive applications, but a minor variant that avoids an extra model call at the end can address that without losing quality. 

\paragraph{Comparison to large models} Gemini 1.5 pro costs more than 30x per token than Gemini 1.5 Flash 8B - so even if the decomposed FT adds 2-3x cost over the base CoT on a small model, it is still much cheaper to run the whole process on Flash 8B than CoT on Gemini pro. Similarly, Gemini 2.5 pro is approximately 5x slower per output token and 100x slower in time to first answer token than 2.5 Flash Lite, so the additional latency that comes from using decomposed FT is much smaller than the additional latency that would come from using a large model.

\section{Discussion}
Our study utilized datasets designed for automation to tackle the challenge of user intent identification, despite their inherent limitations such as noise and information gaps. We observe that fine-tuning alone does not surpass Chain-of-Thought, especially in noisy data scenarios. However, our two stage decomposition exhibited superior performance delivering significant improvements regardless of data quality. This improvement can be attributed to the cleaning process and the combination of prompts and fine-tuning, which effectively mitigated the impact of data noise.

Furthermore, our approach significantly reduced the storage footprint of individual screenshots by summarizing each screen independently, thereby minimizing the required tokens for representation. This reduction in token usage is particularly beneficial for on-device models with limited context windows, enabling them to handle longer trajectories more effectively.

\section{Ethical Considerations \& Risks}
Autonomous agents offer significant innovation, but their development necessitates careful ethical consideration, particularly regarding user privacy. Our research, which aims to interpret user intent from UI interactions, inherently involves sensitive data. We particularly study small models that can run on-device, thereby reducing some of the privacy risks associated with transmitting data to external servers. Furthermore, accurately understanding user intents can greatly benefit users through enhanced personalization, improved work efficiency, and facilitating future recall of past activities on their devices. 
While this work focuses on intent understanding, the development of agents capable of autonomously completing actions requires extreme care. The potential for for misalignment with user intentions and the need for robust safeguards must be thoroughly addressed to ensure responsible deployment.

\section{Limitations}
We acknowledge several discrepancies between our datasets and real-world user behavior. The datasets predominantly feature English-language, U.S.-centric web interactions, restricting our analysis to this specific demographic. In contrast, real-world users frequently navigate multiple applications, adapt their goals on the fly, and exhibit varying levels of digital literacy, resulting in more complex and unpredictable interaction patterns. The Mind2Web dataset's single-website limitation further deviates from the multi-site nature of typical user tasks. Additionally, our study's reliance on Android and web environments limits the generalizability of our findings to other platforms.

\typeout{}
\bibliography{custom}

\appendix
\section{Preprocessing Details}
\label{appendix:preprocessing_details}

The data preprocessing pipeline for the Mind2Web and AndroidControl datasets involved several tailored steps. For image preprocessing, we adopted a holistic approach, where entire screenshots were resized as necessary to conform to model input specifications, deliberately avoiding patch-based methods. For the Mind2Web dataset, full-webpage screenshots were first processed by cropping them to a uniform size of 1280×768. This crop was specifically defined by the bounding box of the user's interaction, ensuring this critical area dictating the action was captured within random margins before the image was resized. In contrast, the AndroidControl dataset, with its uniformly sized mobile screenshots 1080×2400, only necessitated resizing and the subsequent visual overlaying of a bounding box derived from the available user action coordinates. The action extraction process also varied: Mind2Web provided action details directly (which includes information like bounding box coordinates of the target element), whereas for AndroidControl, it was necessary to identify the interacted UI element using its coordinates (which define its bounding box) and then retrieve its name via the accessibility tree. Note that for Gemini experiments on Mind2Web, we additionally filtered examples by domain name to comply with Google-extended policy\footnote{https://blog.google/technology/ai/an-update-on-web-publisher-controls/}. The resulting test set size was reduced from 1005 to 681.

For the gold standard extracted goal, we use the high-level goal for each dataset. As mentioned in Section~\ref{sec:background:datasets}, the annotation process of AndroidControl was less rigorous than that of Mind2Web, resulting in noisier labels. Furthermore, AndroidControl labels, designed to simulate real user instructions, often contain irrelevant information that cannot be inferred from the trajectory (e.g., ``I’m hungry, order an olive pizza from DoorDash''). To mitigate the impact of this noise, we cleaned the labels using Gemini 1.5 Pro (Prompt in \ref{fig:cleaning_prompts}). This cleaning still doesn't completely provide clean goals like Mind2Web’s validation process. We find that even after applying a prompt-based cleaning, manual validation on 100 examples (following the annotation protocol in~\citet{berkovitch2024identifying}) makes changes to $\sim$ 30\% of the label intents. 

Finally, a common preprocessing step was applied to the goals from both Mind2Web and AndroidControl. We noted that the specific application or website name was often available with the interaction data. Yet, this platform-specific information is not directly or consistently inferable from the visual input of screenshots and the action sequences alone. To address this, we programmatically isolated these platform identifiers from the core user intent, restructuring the label into an "app-name/website; intent" format. This approach serves a dual purpose: it retains platform information, useful for contextual fine-tuning, and also allows for the simple removal of this identifier prior to evaluation. Such removal ensures that our assessment accurately reflects the model's capability to interpret user intent, rather than its ability to identify the specific platform. As a result, any distorting effects from platform recognition on the evaluation metrics are prevented, which is important given that platform identification is not a primary focus of this study.

\section{Fine-Tuning Details}
\label{appendix:finetuning_details}
For the fine-tuning process, we adapted slightly distinct approaches for the Gemini and Qwen models, largely adhering to established practices. 

The Gemini models were fine-tuned following procedures analogous to those described described at~\url{https://ai.google.dev/gemini-api/docs/model-tuning}. A learning rate of $1e-6$ was used without specific hyperparameter tuning, and a batch size of 16 was employed. Training proceeded for a maximum of two epochs, with checkpoints saved at intervals of 20 steps. The Gemini model chosen was the one that achieved the minimum negative log-likelihood on its respective validation data, effectively employing an early stopping strategy based on this metric. Similarly, for the Qwen2-VL-7B model, we followed methodology outlined in the Hugging Face VL fine-tuning cookbook\footnote{\url{https://huggingface.co/learn/cookbook/en/fine_tuning_vlm_trl}}. This included adopting the author's recommended hyper-parameters, such as the default learning rate of $2e-4$. Due to memory constraints, a batch size of 1 was employed for Qwen. Training was also conducted for a maximum of two epochs, and checkpoints were saved every 20 steps, as suggested in the tutorial. Consistent with the Gemini models, the final Qwen model was selected to minimize negative log-likelihood on validation data.
For the AndroidControl dataset, we used 5,000 training examples and 137 validation examples randomly sampled from the train set. For Mind2Web we used 900 training examples and 90 validation examples. 

\section{Expert Annotation Labeling}
\label{appendix:expert_labeling}
Table~\ref{tab:detailed_annotated_vs_gold} expands on the results shown in Table \ref{tab:expert_labels} of the main text, providing the detailed BiFact precision and recall scores in addition to the F1 score for the comparison against expert annotations and original dataset labels on the AndroidControl dataset. As is evident from these numbers, the increase in recall when evaluating against expert labels is particularly significant. This suggests that many of
the facts included in the original dataset labels were not actually fulfilled (or were unfulfillable) within the recorded user interaction trajectories. Consequently, the higher recall achieved against the expert-annotated labels more accurately reflects the model's performance on verifiable and achievable intents. 

\begin{table*}
\small
\begin{tabular}{l|lll|lll}
\toprule
& \multicolumn{3}{c|}{Expert Labels} & \multicolumn{3}{c}{Dataset Labels} \\
& F1 & Precision & Recall & F1 & Precision & Recall \\
\midrule
Gemini Flash 8B \\ 
\midrule
CoT & 0.652 & 	0.674 &	0.714 &	0.580 &	0.600 &	0.663 \\
E2E-FT & 0.590 &	0.636 &	0.623 &	0.565 &	0.626 &	0.601 \\
Decomposed-FT & 0.701	& 0.714 & 	0.762&	0.596&	0.639 &	0.655  \\
\midrule
Gemini-1.5-Pro \\ 
\midrule 
CoT  & 0.724 &	0.688 &	0.862 &	0.635 &	0.617 &	0.746   \\
\bottomrule
\end{tabular}
\caption{\label{tab:detailed_annotated_vs_gold} A comparison of BiFact F1, precision and recall scores for intent prediction on the AndroidControl dataset, using expert annotations and dataset labels as ground truth. }
\end{table*}

\begin{table*}[htbp]
\small
    \centering
    \begin{tabularx}{\textwidth}{l|XXX|X|XXX|X}
\toprule
& \multicolumn{4}{c}{\textbf{Mind2Web}} & \multicolumn{4}{c}{\textbf{AndroidControl}} \\
\midrule
& \multicolumn{3}{c|}{BiFact} & \multicolumn{1}{c|}{Bi-NLI}  & \multicolumn{3}{c|}{BiFact} & \multicolumn{1}{c}{Bi-NLI} \\
Method & F1 & Precision & Recall  & &  F1 & Precision & Recall &  \\
\midrule
Gemini Flash 8B  \\
\midrule
CoT & 0.660 & \textbf{0.751} & 0.656 & 0.326 & \textbf{0.594} & \textbf{0.628} & 0.660 & 0.302 \\
Decomposed-non-FT & \textbf{0.718} & 0.717 & \textbf{0.792} & 0.221 &  0.528 & 0.488 & {\bf 0.719} & 0.185  \\
\midrule 
Gemini-1.5-Pro  \\
\midrule
CoT & 0.721 & \textbf{0.761} & 0.740 &  0.331 &  \textbf{0.634} & \textbf{0.612} & 0.767 & 0.347  \\
Decomposed-non-FT & \textbf{0.732} & 0.700 & \textbf{0.859} & 0.213& 0.512 & 0.441 & {\bf 0.791}  & 0.228 \\
\bottomrule
\end{tabularx}
    \caption{BiFact results on the Mind2Web and AndroidControl datasets using Gemini 1.5 Flash 8B and Gemini 1.5 Pro.}
    \label{tab:non_ft}
\end{table*}

\begin{table*}
\small
\begin{tabular}{l|lll|lll|lll}
\toprule
& \multicolumn{3}{c|}{DOMAIN UNSEEN} & \multicolumn{3}{c|}{TASK UNSEEN} & \multicolumn{3}{c}{WEBSITE UNSEEN} \\
& F1 & Prec. & Recall & F1 & Prec. & Recall & F1 & Prec. & Recall \\
\midrule
Gemini Flash 8B \\ 
\midrule
CoT & 0.656 & 0.767 & 0.641 & 0.651 & 0.717 & 0.655 & 0.692 & 0.786 & 0.664 \\
E2E-FT & 0.665 & 0.686 & 0.686 & 0.606 & 0.631 & 0.618 & 0.674 & 0.618 & 0.694 \\
Decomposed-FT & \textbf{0.747} & \textbf{0.800} & \textbf{0.752} & \textbf{0.732} & \textbf{0.823} & \textbf{0.710} & \textbf{0.817} & \textbf{0.899} & \textbf{0.785} \\
\midrule
Gemini-1.5-Pro \\ 
\midrule 
CoT  & 0.723 &	0.774 &	0.734 &	0.731 &	0.762 &	0.762 &	0.756 &	0.785 &	0.774  \\
\bottomrule
\end{tabular}
\caption{\label{tab:detailed_test_mind2web} Detailed BiFact-based performance breakdown on different subsets of the Mind2Web test set. }
\end{table*}

\section{Human Preference Annotation}
\label{appendix:human_preference}
We presented the rater with a full trajectory of screenshots and actions, and then asked the following question:
``After you have seen the trajectory, which intent better describes the trajectory? A or B.'' The choices A and B contained
either CoT or Decomposed-FT. The order of the two options were randomized in each question and the names of the methods were not shown to the respondent.
The decoding of choices to model name was only done after the rater had finished the task.

\section{Results Using an Alternative LLM for Evaluation}
\label{appendix:alternative_llm}
To demonstrate that our findings are robust to the choice of the underlying model for our evaluation metric, we present a replication of our main results in Table \ref{tab:appendix_bifact_flash}. In this analysis, we replaced the LLM used by BiFact, substituting Gemini Pro with Gemini Flash.

The results show that the overall performance trends and the relative ranking of the evaluated models remain consistent, confirming that our main conclusions are not dependent on a specific evaluation LLM. 
\begin{table*}[htbp]
    \centering
    {\small
    \begin{tabularx}{\textwidth}{l|XXX|XXX}
\toprule

& \multicolumn{3}{c}{\textbf{Mind2Web}} & \multicolumn{3}{c}{\textbf{AndroidControl}} \\
\midrule
& \multicolumn{3}{c|}{BiFact} & \multicolumn{3}{c}{BiFact}  \\
Method & F1 & Precision & Recall & F1 & Precision & Recall \\
\midrule
Gemini Flash 8B  \\
\midrule
CoT & 0.681 & 0.796 & 0.656 & 0.623 & 0.675 & 0.681 \\
E2E-FT & 0.678 & 0.726 & 0.680 & 0.651 & 0.72 & 0.679 \\
Decomposed-FT & \textbf{0.794} & \textbf{0.874} & \textbf{0.771} &  \textbf{0.685} & \textbf{0.741} & \textbf{0.718} \\ 
\midrule 
Gemini-1.5-Pro  \\
\midrule
CoT & 0.746 & 0.805 & 0.757 & 0.701 & 0.695 & 0.811 \\
\bottomrule
\end{tabularx}
}
    \caption{Replication of Main Results (Table~\ref{tab:gemini_mind2web}) Using an Alternate LLM. BiFact results on the Mind2Web and AndroidControl datasets, using Gemini-Flash as the underlying LLM for BiFact rather than Gemini-Pro. This demonstrates that our main findings are robust to the choice of the underlying LLM used for evaluation.}

    \label{tab:appendix_bifact_flash}
\end{table*}

\section{Detailed Test Set Performance Breakdown}
\label{appendix:test_set_breakdown}
The test-sets for Mind2Web~\citep{deng2024mind2web} and AndroidControl~\citep{li2024effects} have multiple types of unseen data. In this appendix, we provide a more detailed breakdown of the BiFact performance scores on each of the subsets of the test sets.

The detailed performance breakdown on the Mind2Web test set, as presented in Table \ref{tab:detailed_test_mind2web}, offers several key insights into model generalization. As might be expected, the standard end-to-end fine-tuned (E2E-FT) model using Gemini Flash 8B performs worse than the COT approach on data from previously unseen tasks (TASK UNSEEN) and unseen websites (WEBSITE UNSEEN). However, its performance is notably on par with the COT model in the DOMAIN UNSEEN category. This pattern suggests that while the E2E-FT model may be tuned somewhat towards specific tasks and characteristics of websites present in its training data, its ability to handle completely new types of tasks at a broader domain level is not further compromised compared to the prompt-based COT method.
In stark contrast, the Decomposed-FT model (Gemini Flash 8B) demonstrates strong performance, consistently outperforming both the COT and E2E-FT methods across all three challenging unseen categories (DOMAIN, TASK, and WEBSITE UNSEEN). Furthermore, its performance in these generalization scenarios surpasses that of the larger Gemini 1.5 Pro (COT) model, particularly on unseen domains and websites. This robust performance can be attributed to the sophisticated fine-tuning scheme employed for the Decomposed-FT model. This comprehensive approach—which involves using prompts for structured interaction summarization, meticulous data cleaning through label refinement, and then fine-tuning on these processed, higher-quality inputs—makes the model significantly less vulnerable to common problems associated with regular fine-tuning, such as overfitting to training set specifics or sensitivity to label noise. 
The primary limitation highlighted by Table \ref{tab:detailed_test_mind2web}, when comparing the Decomposed-FT model (using Gemini Flash 8B) to the Gemini-1.5-Pro model, is its slightly lower recall in the TASK UNSEEN category (0.710 for Decomposed-FT vs. 0.762 for Pro). This specific gap suggests that while highly effective, the training scheme could potentially benefit from exposure to a more diverse range of task examples to further enhance generalization for entirely novel tasks, even when encountered within familiar website or domain contexts.

\section{Detailed Analysis of the non-finetuned Decomposed ablation}
\label{appendix:test_decomposed_analysis}

To provide a complete picture and address potential inquiries regarding the performance of our approach without the crucial fine-tuning of the intent extraction stage, this section offers a more in-depth analysis of the non-finetuned ablation of our decomposed method. 
Table \ref{tab:non_ft} presents a comparison of the non-finetuned decomposed ablation with the CoT baseline. Since neither method requires fine-tuning, we can demonstrate their performance across both small and large models. 
Notably, our full method, Decomposed-FT, surpasses both the non-finetuned decomposed variant, as evidenced in the ablation study in Table \ref{tab:ablation}, and the CoT baseline, as shown in Table \ref{tab:gemini_mind2web}. 
The results in Table \ref{tab:non_ft} indicate that the non-finetuned decomposed method demonstrates strong performance on the Mind2Web dataset, yet underperforms considerably compared to the CoT baseline on the Android Control dataset, a trend consistent across both small and large model sizes. This performance differential can be attributed to the inherent verbosity of the non-finetuned decomposed method, which generates a higher average number of atomic facts per predicted intent compared to CoT. Specifically, for Android Control, the non-finetuned decomposed approach produced an average of 4.0 facts versus 2.5 for CoT, while gold has 3.0 facts. For Mind2Web, these figures were 4.4 for Decomposed versus 2.8 for CoT, while the gold  has 4.4 facts. This increased verbosity correlates with the observed lower precision of the non-finetuned decomposed method across both datasets. Conversely, it also correlates with the superior fact-level performance (BiFact), of the decomposed method on Mind2Web, where the gold annotations are themselves more verbose than those for Android Control. As our main results show, subsequent fine-tuning of the second-stage model demonstrably improves precision by training the model to selectively include only the most relevant facts in the final intent formulation, leading to the significantly better performance of our proposed method.

\section{Comparison with SummAct }
\label{appendix:compwithsummact}

In SummAct \citep{zhang2024summact}, each input interaction is represented as a short textual description of the specific UI element with which the user interacted and the respective user action. Intent extraction is then performed hierarchically, in two steps. First, the sequence of interactions is summarized into a shorter sequence of mid-level sub-goals, using few-shot prompting. Then, the sequence of sub-goals is further summarized into the final high-level intent description, using a model that is finetuned to produce the gold intent given the output of the first step. While this method seems somewhat similar to ours in that it decomposes intent extraction into two subsequent steps, the two methods differ substantially, in both their input and their decomposed steps.
SummAct considers only localized textual input, describing just the UI element with which the user interacted, while ignoring the full screen and its visual layout. This restricts the model from understanding the wider context, like the elements the user didn’t choose to interact with, or visual cues that may influence user behavior. In contrast, our model, considers the full screenshot information. Thus, in our first step, the model generates a textual summary for each interaction step, which considers the broader screen context. Subsequently, our second step directly summarizes these interaction level descriptions into the final high-level intent description, using a sophisticated fine-tuning approach that synchronizes the inputs for this step with the gold output. This method does not require an intermediate step of generating sub-goal descriptions, like SummAct. Further, while SummAct's finetuned model required intervention with the attention mechanism to perform well, which may not be accessible or practical in various settings, our finetuning approach allowed us to finetune available models as is, without further intervention. 
A potential advantage of SummAct’s text-only approach may be computational efficiency as it consumes smaller inputs and uses fewer model calls.

\section{Prompts}
\label{appendix:prompts}

\setcounter{table}{0}
\setcounter{figure}{0}
\renewcommand{\thetable}{I\arabic{table}}
\renewcommand{\thefigure}{I\arabic{figure}}

\hspace*{5cm}
\begin{figure*}[p]
    \centering
    \fbox{\includegraphics[width=.8\textwidth]{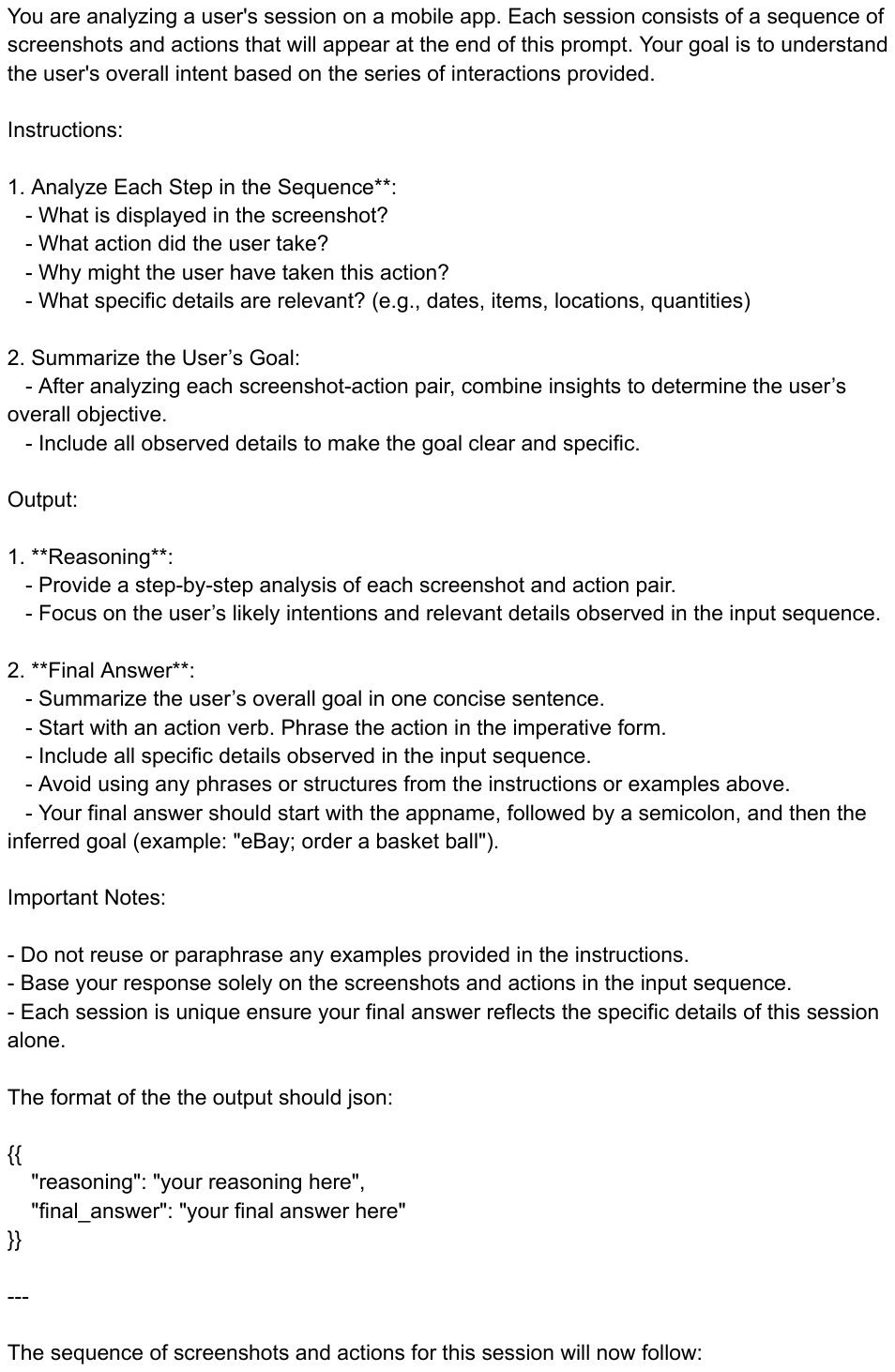}}
    \caption{CoT model prompt, used as the baseline as described in Section~\ref{sec:baselines}}
    \label{fig:prompt_cot}
\end{figure*}

\begin{figure*}
    \centering
    \fbox{\includegraphics[width=.8\textwidth]{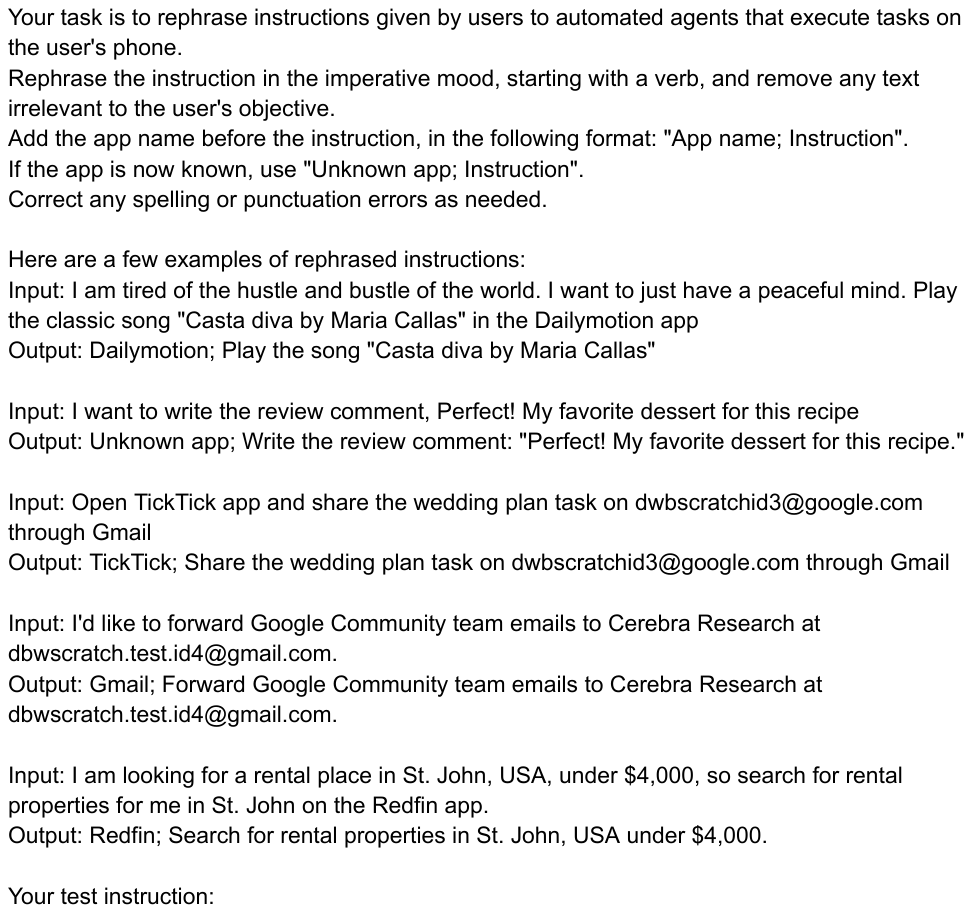}}
    \caption{AndroidControl cleaning prompt, used to automatically clean the dataset as described in Section~\ref{sec:datasets_preprocessing}}
    \label{fig:cleaning_prompts}
\end{figure*}

\begin{figure*}
    \centering
    \fbox{\includegraphics[width=.8\textwidth]{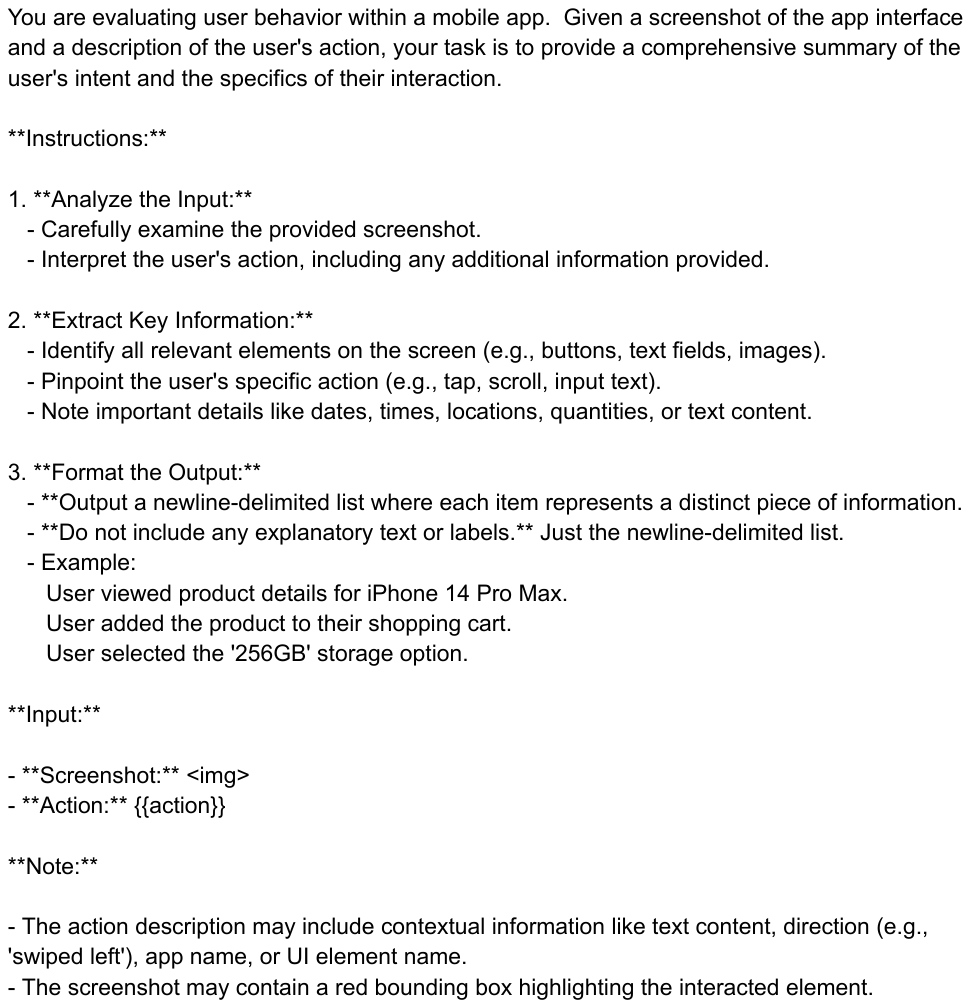}}
    \caption{Interaction summarization prompt, used to summarize single screen interaction as explained Section~\ref{sec:interaction_summarization}}
    \label{fig:summarisation_prompt}
\end{figure*}

\begin{figure*}
    \centering
    \fbox{\includegraphics[width=.8\textwidth]{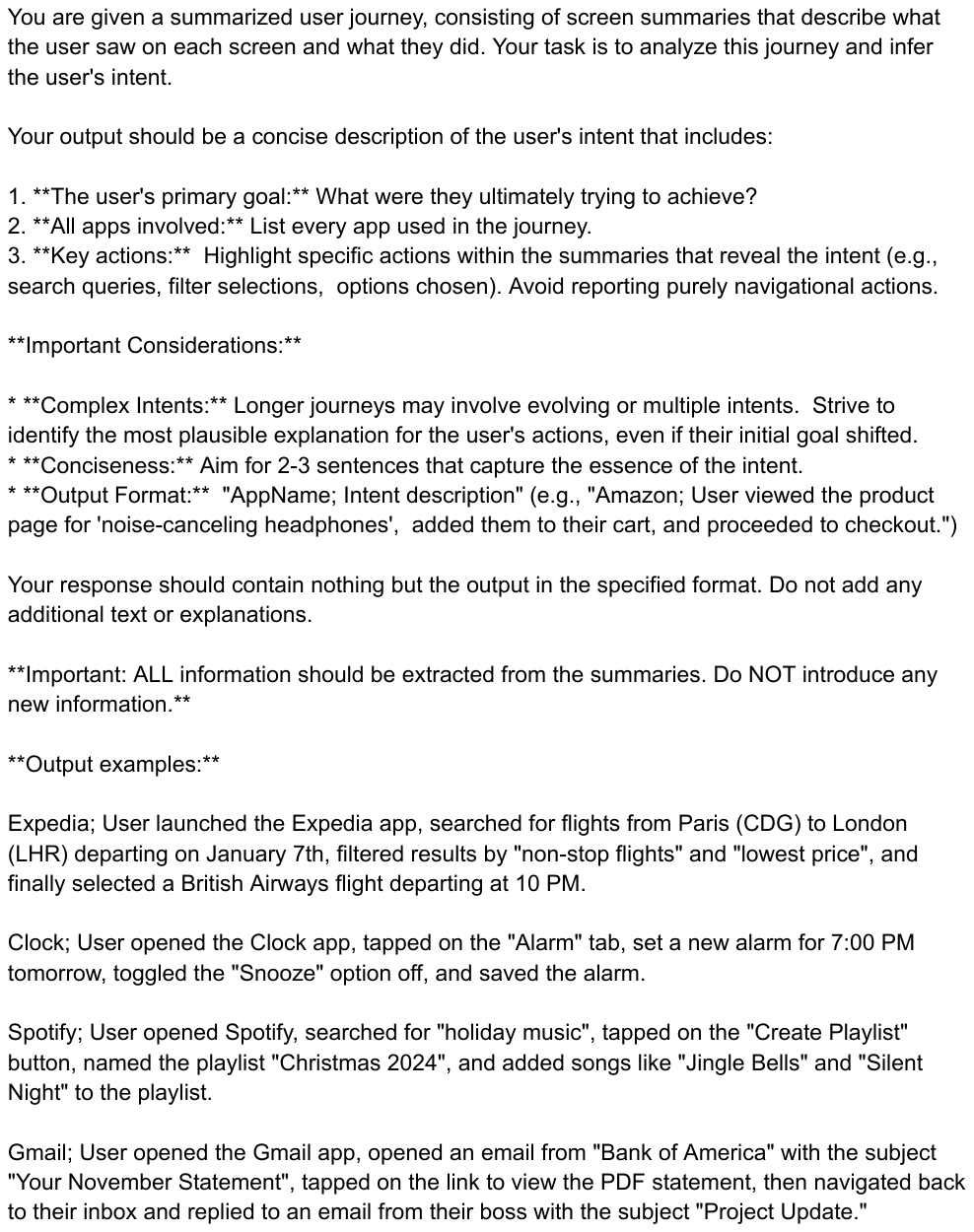}}
    \caption{Session-level intent prompt, used to fuse single interaction summarise to a single intent as explained in Section~\ref{sec:session_Level_Intent}}
    \label{fig:mobile_extraction_prompt}
\end{figure*}

\begin{figure*}
    \centering
    \fbox{\includegraphics[width=.8\textwidth]{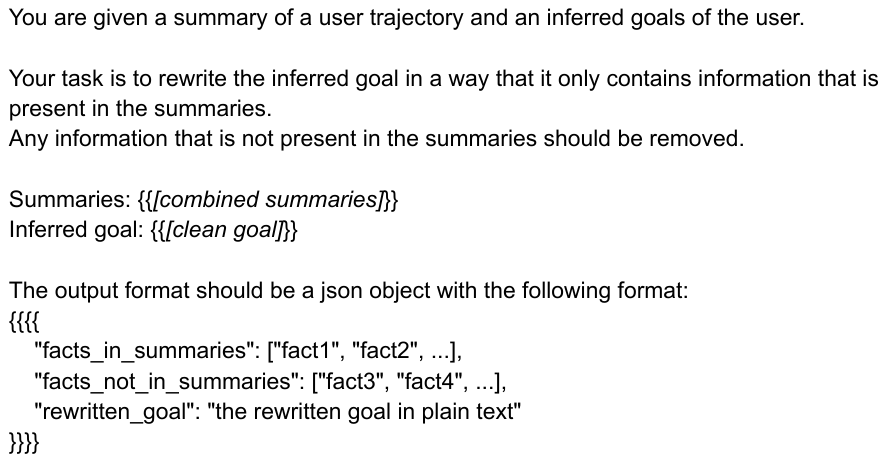}}
    \caption{Label refinement prompt, used to refine the label prior to the fine-tunning step as explained in Section~\ref{sec:session_Level_Intent}}
    \label{fig:label_refinement_prompt}
\end{figure*}

\end{document}